\documentclass[10pt, conference, letterpaper]{IEEEtran}
\usepackage[utf8]{inputenc}
\usepackage{xspace}
\usepackage{subfig}
\usepackage[font=small]{caption}

\usepackage{xspace} 
\usepackage{multirow}
\usepackage{mathtools}
\usepackage{units}
\usepackage{todonotes}
\usepackage{comment}

\usepackage{amssymb}
\usepackage{booktabs}

\usepackage[colorlinks]{hyperref}
\hypersetup{
    citecolor = {magenta},
    linkcolor = {black}
}

\usepackage{bbding}

\usepackage{pifont}

\newcommand{\DEF}{\stackrel{\text{def}}{=}}
\newcommand{\aka}{\emph{a.k.a.}\xspace}
\newcommand{\modelbase}{$\mathsf{M}_{source}$\xspace}
\newcommand{\modeltarget}{$\mathsf{M}_{target}$\xspace}
\newcommand{\taskbase}{$\mathcal{T}_{source}$\xspace}
\newcommand{\tasktarget}{$\mathcal{T}_{target}$\xspace}

\usepackage{wasysym}
\newcommand{\circleempty}{\textcolor{gray}{\Circle}}
\newcommand{\circlefull}{\textcolor{gray}{\CIRCLE}}
\newcommand{\circlesemi}{\textcolor{gray}{\LEFTcircle}}
\newcommand{\tinytiny}[1]{\fontsize{5.5}{10}\selectfont \textcolor{gray}{$\,\pm\,#1$}}
\newcommand{\tinytinyalternate}[1]{\fontsize{7}{10}\selectfont \textcolor{gray}{$\,\pm\,#1$}}

\newcommand{\mirage}{$\mathtt{MIRAGE19}$\xspace}
\newcommand{\appclassnet}{$\mathtt{AppClassNet}$\xspace}

\newcommand{\proto}{$\mathtt{ProtoNet}$\xspace}
\newcommand{\maml}{$\mathtt{MAML}$\xspace}
\newcommand{\relation}{$\mathtt{RelationNet}$\xspace}

\newcommand{\BESTPERF}{\cellcolor{green!20!white}}
\newcommand{\WORSTPERF}{\cellcolor{red!8!white}}

\newcommand{\base}{$\mathtt{Baseline}$\xspace}
\newcommand{\basetl}{$\mathtt{Baseline\text{-}TL}$\xspace}
\newcommand{\basesimclr}{$\mathtt{SimCLR}$\xspace}
\newcommand{\basesupcon}{$\mathtt{SupCon}$\xspace}
\newcommand{\basepp}{$\mathtt{Baseline(ClassEmb)}$\xspace}
\newcommand{\basepptl}{$\mathtt{Baseline(ClassEmb)\text{-}TL}$\xspace}
\newcommand{\baseppsimclr}{$\mathtt{SimCLR(ClassEmb)}\xspace$}
\newcommand{\baseppsupcon}{$\mathtt{SupCon(ClassEmb)}\xspace$}
\newcommand{\rfsslr}{$\mathtt{Baseline(LR)}$\xspace}
\newcommand{\rfssnn}{$\mathtt{Baseline(NN)}$\xspace}
\newcommand{\rfsdlr}{$\mathtt{RFS\text{-}distill(LR)}$\xspace}
\newcommand{\rfsdnn}{$\mathtt{RFS\text{-}distill(NN)}$\xspace}
\newcommand{\simclr}{$\mathtt{SimCLR}$\xspace}

\newcommand{\supcon}{$\mathtt{SupCon}$\xspace}
\newcommand{\RF}{$\mathtt{RF}$\xspace}
\newcommand{\CNNtwo}{$\mathtt{CNN\text{-}2}$\xspace}
\newcommand{\CNNfour}{$\mathtt{CNN\text{-}4}$\xspace}

\usepackage{colortbl}
\usepackage[square,sort,comma,numbers]{natbib}

\begin{document}

\bstctlcite{IEEEexample:BSTcontrol} 
\title{
Many or Few Samples? \\
\Large{Comparing Transfer, Contrastive and Meta-Learning in Encrypted Traffic Classification}
}

\author{
\IEEEauthorblockN{Idio Guarino$^\dagger$, Chao Wang, Alessandro Finamore, Antonio Pescap\`e$^\dagger$, Dario Rossi}
\IEEEauthorblockA{$\dagger$ Universit\'{a} degli Studi di Napoli Federico II, Huawei Technology France}
}

\newcommand\eg{\emph{e.g.,}\xspace}
\newcommand\ie{\emph{i.e.,}\xspace}

\maketitle

\begin{abstract}
The popularity of Deep Learning (DL),
coupled with network traffic visibility reduction
due to the increased adoption of HTTPS, QUIC, and DNS-SEC,
re-ignited interest towards Traffic Classification (TC).
However, to tame the dependency from task-specific large 
labeled datasets we need to find better ways to 
learn representations that are valid across tasks.
In this work we investigate
this problem comparing \emph{transfer learning},
\emph{meta-learning} and \emph{contrastive learning}
against reference Machine Learning (ML) tree-based
and  monolithic DL
models (16 methods total). Using two 
publicly available datasets, namely \mirage (40 classes)
and \appclassnet (500 classes), we show that 
($i$) by using DL methods on large datasets we can obtain more general representations
with ($ii$) contrastive learning methods yielding the best performance
and ($iii$) meta-learning the worst one. While
($iv$) tree-based models can be impractical for large tasks 
but fit well small tasks, ($v$) DL methods that reuse better learned representations
are closing their performance gap against trees also for small tasks.
\end{abstract}



\section{Introduction}
\label{sec:intro}

Network monitoring is paramount for networks operation,
with 
\emph{Traffic Classification} (TC) being a strategic activity to empower
better decision making. 
Started more than a decade ago, 
the transition towards completely encrypted network traffic 
is today reinvigorated by the growing adoption of
QUIC~\cite{ruth2018first-quic}, DNSSEC~\cite{roth2019tracking-dnssec}, or initiatives like Apple 
iCloud Private Relay~\cite{trevisan2023measuring-icloudprivaterelay}---encryption is here to stay for 
end-users benefit at the cost of reduced traffic visibility for network~operators.

For recovering from such reduction and pursuing 
better network management and automation,
both academia and industry started the quest for smarter TC solutions. 
A first wave of Machine Learning (ML)-based solutions
was already introduced twenty years ago, while more recent
proposals focus on Deep Learning (DL)~\cite{yang2021tnsm-zero-day}---nowadays,
TC is the king 
of supervised modeling tasks in network traffic analysis.

Training and managing ML/DL models for networks
faces the ``infinite loop'' of collecting new
data and re-train models to keep them up to date.
The key to break this cycle resides in 
($i$) training more generalized models and 
($ii$) adapting them to new scenarios by means of little-to-no extra data. 
This calls for exploring 
\emph{transfer learning} and \emph{Few-Shot Learning} (FSL), two
DL techniques designed to foster better representation learning.
These techniques allow to reuse what
learned from a source task $\mathcal{T}_{source}$ 
to address a different task $\mathcal{T}_{target}$.
Within the same scope, \emph{contrastive learning}
and \emph{self-supervision} come with the promise of better knowledge extraction thanks to smarter use of data augmentation.

Considering TC literature, we find several studies 
investigating FSL in the context of intrusion 
detection~\cite{ouyang2021tc-fs-ids,
rong2021ijcnn-umvd-fsl-unseen-malware,
yu2020ieeeaccess,
liang2022-toii-oics-vfsl,
zheng2020www-rbrn,
xu2020tifs-fcnet,
zhao2022ieeecomlet-festic,
feng2021-ifipnetworking-fcad}.
Yet, ($i$) most of them do not follow the conventional
meta-learning training protocol, hence (arguably) biasing
their takeaways (see Sec.~\ref{sec:background}, Sec.~\ref{sec:related-work});
only \cite{horiwicz2022imc-fsl-contrastive-learning,ziyi-networking22-cl-etc} investigated
contrastive learning; no previous literature relies on state-of-the-art transfer learning techniques~\cite{chen2019icml-baseline,tian2020eccv-rethingkingfewshot}.
Moreover, ($ii$) little attention
has been spent on investigating ML tree-based approaches.
Considering datasets, ($iii$) we find most studies using only up to 20 classes,
i.e., possibly not enough variety to understand 
the training methodologies pros and cons.
Last, ($iv$) most methods rely on payload bytes, reshaped as large 2d matrix, 
which can be costly to track for monitoring systems.

Motivated by the previous considerations, in this work we study
transfer learning, meta-learning, and contrastive learning
(a total of 16 variants) by means of two public
datasets, namely \mirage (40 classes)
and \appclassnet (500 classes). Beside benchmarking the methods,
we aim at 
\emph{understanding up to which extent 
large TC datasets can be used for learning
better representation via DL.}
For completeness, we study also ML tree-based models
and traditional CNN-based models.

Our results show that TC literature might have overstated 
meta-learning methods benefits which are the worst performing
in our assessment (at least $-14/18$\% from
the best alternative). 
Conversely, and aligned with
recent research in computer vision, 
contrastive learning (especially in a supervised setting) is quite effective,
yet suffering from computational costs. 
Tree-based models
are still superior to all methods but, while they can
grow too deep---a $500$ classes model on \appclassnet can
grow up to a depth of $117$ corresponding to $416$GB file size---they are still the most practical
solution when modeling classes with $\leq$$1,\!000$ samples/class.
Still, the best DL alternatives are closing the performance
gap against tree-based models.

In the following, we start by introducing the principles behind 
the DL techniques (Sec.~\ref{sec:background})
and reviewing related computer vision and TC literature (Sec.~\ref{sec:related-work}).
We continue stating our research goals (Sec.~\ref{sec:research-questions})
used when designing our experiments (Sec.~\ref{sec:methodology}).
We conclude by discussing our results (Sec.~\ref{sec:evaluation})
and outlining future research avenues (Sec.\ref{sec:conclusion}).


\begin{figure*}
\includegraphics[width=\textwidth]{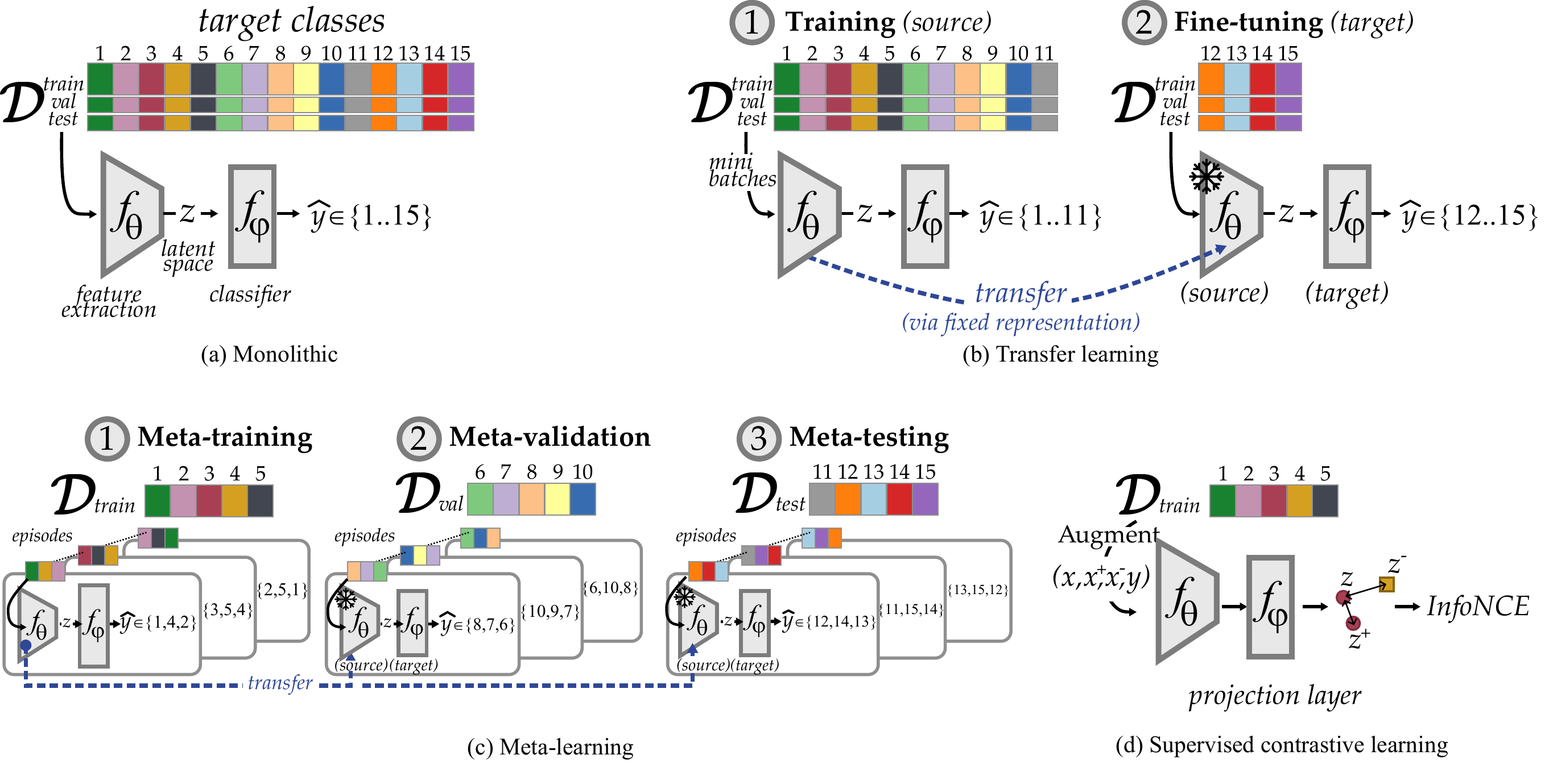}
\caption{DL training strategies comparison.}
\label{fig:sketch-training}
\end{figure*}

\section{Background}
\label{sec:background}

In this section, we review DL models training principles for
monolithic, transfer, meta- and contrastive learning.
We focus on a supervised task $\mathcal{T}$ 
with a dataset $\mathcal{D} = \{(x_1, y_1) \cdots (x_m, y_m)\}$, 
where each input sample $x \in \mathcal{X}$ maps to a class label $y \in \mathcal{C}=\{1...N\}$. $\mathcal{D}$ is partitioned into $\mathcal{D}_{train}$, 
$\mathcal{D}_{val}$ and $\mathcal{D}_{test}$ and each ($x_i$, $y_i$) belongs
to one partition only.

\subsection{Monolithic training}

A DL model embodies a 
function $f_{\theta,\varphi} : \mathcal{X} \rightarrow \mathcal{C}$ mapping input samples to labels.
As sketched in Figure~\ref{fig:sketch-training}(a), 
the model decouples 
feature extraction (\aka encoding, embedding) $f_\theta : \mathcal{X} \rightarrow \mathcal{Z}$ 
 from classification $f_\varphi : \mathcal{Z} \rightarrow \mathcal{C}$
 by means of an intermediate latent space $\mathcal{Z}$ where each class projected  samples should be
easily separable. 
Training such model requires tuning the function parameters 
$(\theta,\varphi)$ to minimize the difference (\aka the classification loss)
between true labels $y$ and predicted ones $\hat{y}$. 
To do so, during training $\mathcal{D}_{train}$ is re-played multiple 
times (\aka epochs) through the model by means of mini-batches 
(each being a collection of inputs sampled from $\mathcal{D}_{train}$) 
to progressively adjust the parameters. Each training epoch is usually validated by assessing classification performance on $\mathcal{D}_{val}$, 
with model parameters temporarily frozen for quantifying over-fitting (i.e., verifying that the model  can classify data not belonging to the training set)
and searching optimal hyper-parameters. 
Lastly, testing on $\mathcal{D}_{test}$ yields the final classification performance.
As sketched in Fig.~\ref{fig:sketch-training}(a), 
training, validation, and testing ($i$) share the same task
which targets all available classes---the model is \emph{monolithic}---and ($ii$) the dataset partitions
are scanned at least once.\footnote{In each training epoch, the dataset
is shuffled and then sequential micro-batches are formed---this 
is semantically a scan of the dataset. The process is the same
in validation and testing, although shuffling is unnecessary.}
As we shall see, while transfer learning respects both principles,
meta-learning violates them both.

\subsection{Transfer learning}
\label{sec:background-transfer-learning}

Transfer learning~\cite{weiss2016-survey-transfer-learning} 
enables to address a target task \tasktarget via
a model \modelbase trained for a different task \taskbase. 
Intuitively, the stronger
the relationship between \taskbase and \tasktarget,
the higher the chance the knowledge in \modelbase can be
used (hence transferred) to \tasktarget.
However, exactly quantifying task affinity 
is challenging~\cite{crawshaw2020arxiv-survey-multitask-learning},
but training on large datasets is a ``cheap'' way to bypass
the problem~\cite{sun2017iccv-revisiting-large-datasets}.
As sketched in Fig.~\ref{fig:sketch-training}(b), \modeltarget is composed of
the already trained feature extractor $f_\theta^{(source)}$ (\aka, trunk, backbone, encoder)
and a new classifier $f_\varphi^{(target)}$ to be trained from scratch
and possibly having a different number of classes
and/or architecture with respect to \taskbase.
Given the mixture of trained and untrained parameters,
training of \modeltarget is referred to as \emph{fine-tuning}
and can be performed in a few ways. Fig.~\ref{fig:sketch-training}(b) shows 
\emph{fixed representation} scenarios where the transferred trunk is \emph{frozen} \SnowflakeChevron\xspace when training 
the new classifier, i.e., only the new classifier parameters are optimized.
Fixed representation is commonly used to compare different
representation learning methods, but alternatively, one can fine-tune all parameters at once, or use a hybrid
policy unfreezing trunk parameters after a certain number of epochs.
No matter the selected option, transfer learning still
adheres to monolithic training---dataset partitions are entirely scanned
and \tasktarget enlists all available
classes, yet those are disjoint with respect to \taskbase.

\subsection{Meta-learning and episodic training}
\label{sec:background-meta-learning}
While transfer learning relies on \emph{implicit} tasks affinity, 
meta-learning is designed to \emph{explicitly} push cross-tasks representation extraction.
To do so, ($i$) $\mathcal{D}_{train}$, $\mathcal{D}_{val}$ and $\mathcal{D}_{test}$
contain disjoint set of classes and ($ii$) training is based on
the generation of synthetic tasks $\mathcal{T}_i$ called \emph{episodes}.
During an epoch, 
rather than completing a scan of $\mathcal{D}_{train}$ by means of random mini-batches, 
synthetic tasks $\mathcal{T}_i$ are created by 
randomly sampling a subset of training classes $\mathcal{C}_i^{(train)}$.
Then, for each class two disjoint sets of samples
are randomly formed to drive the learning, namely \emph{support} $\mathcal{S}_i$ and \emph{query} $\mathcal{Q}_i$.
Episodic training is synonym of Few-Shot Learning (FSL)~\cite{wang2020survey-fsl}
and is also known as \emph{(N-way, S-shot) training}:
an episode contains $N$ classes (ways)
each having two sets of samples, namely $S \DEF |\mathcal{S}_i|$ shots and $Q \DEF |\mathcal{Q}_i|$ query samples acting as a ``micro-training'' and
``micro-validation'' set. 
We highlight that sometimes FSL is broadly referred to 
training (or fine-tuning) with small datasets
but without episodic training (e.g.,\cite{horiwicz2022imc-fsl-contrastive-learning}).
However, in this paper FSL is synonym for meta-learning.

Notice that while episodes are small batches of samples,
by design they differ from monolithic training mini-batches as
they guarantee class balance. In monolithic training, mini-batches
are formed by randomly selecting samples across the dataset hence
($i$) they reflect the dataset class imbalance (if any) 
and ($ii$) even for already balanced dataset, 
there is no guarantee that all target classes to be present in a mini-batch. 
Conversely, in meta-learning, an episode has $(S+Q)$ samples for each of the (randomly selected) $N$ target classes (ways).

Overall, this process is known as meta-training and is complemented 
by the remaining two phases, meta-validation and meta-testing
which still rely on fine-tuning the target models, but
recall that $\mathcal{D}_{train}$, $\mathcal{D}_{val}$ and $\mathcal{D}_{test}$
are disjoint.
More in details, the target task \modeltarget
is composed by transferring the meta-learned source model trunk $f_\theta^{(source)}$ 
and fine-tune a classifier specific for an episode.
In other words, as pictured in Fig.~\ref{fig:sketch-training}(c), the evaluation (being validation
or testing) creates multiple models (one per episode) so
overall performance is an
aggregation of per-episode models performance.
As in monolithic training, meta-validation facilitates both
hyper-parameters tuning and over-fitting assessment,
while meta-testing yields the final performance.
Although not strictly required, it is also common to maintain 
the same \emph{N-ways} for meta-train, meta-validation and
meta-test.

As for transfer learning, it is beneficial to use
a $\mathcal{D}_{train}$ offering many classes (e.g., $\geq100$) 
but episodes are commonly small, i.e., $N$=$\{5, 10\}$ classes each having
$S$=$\{1, 5\}$ support samples and $Q$=$\{10, 50\}$ query 
samples---meta-training is designed to push representation learning
through tasks variety.

\subsection{Contrastive learning}
From a geometrical standpoint, a well trained supervised classifier 
should project input samples in the latent space so to
pull together samples from the same class while distancing
them from other classes samples.
Recall that the common model classifier $f_\varphi(\cdot)$ is a simple linear layer,
hence a traditional classification loss (e.g., cross-entropy loss)
\emph{implicitly} trains the feature extractor to achieve
such geometrical property. 
Contrastive learning~\cite{chen2020icml-simclr} is a technique
to \emph{explicitly} regularize latent space geometry
by means of data augmentation and a ``contrastive game'': 
an input sample $x\in\mathcal{X}$ (\aka anchor) is transformed 
(via rotation, cropping, jittering, blurring, etc.) into 
a \emph{positive sample} $x^+$=$\mathsf{Augment}(x)$
while the remaining available samples in $\mathcal{X}$
can be used as \emph{negative samples}; then 
the training loss is formed to penalize cases
where $(f_\theta(x), f_\theta(x^+))$ are far apart 
and $(f_\theta(x), f_\theta(x^-))$ are too close.
Multiple losses can drive the learning process
with \emph{InfoNCE}~\cite{oord2018representation-infonce},
which maximizes the mutual information between anchor and positive samples,
being the most popular choice.
Overall, contrastive learning differs from SMOTE~\cite{chawla2002smote}, ADASYN~\cite{he2008adasyn} and 
other augmentations (e.g, random gaussian jittering) 
which ``blindly'' add variety to $\mathcal{D}_{train}$ 
but no latent space regularization is explicitly introduced.

Contrastive learning is at the core of
Self-Supervised Learning (SSL) and is defined for
unsupervised tasks---with no extra information, a sample
can only be \emph{self-similar}, i.e., a positive sample
is the transformation of the sample itself---with a
wide design space related to the selection of
transformations, positives and negatives samples~\cite{liu2023-surve-clr}.
As from Fig.~\ref{fig:sketch-training}(d), notice 
the presence of $f_\varphi(\cdot)$ acting as a projection
layer, not as a classifier.
Even supervised variants of contrastive learning (e.g.,~\cite{khosla2020neurips-supcon})
are still effectively unsupervised as labels guide the positive samples selection, not a classification loss.



\section{Related work}
\label{sec:related-work}

In this section, we complement the DL training principles introduced so far with
a review of computer vision and traffic classification literature
as summarized in Table~\ref{tab:sota-cv} and Table~\ref{tab:sota-tc}.

\subsection{Transfer learning}
\label{sec:transfer_learning}

\noindent \textbf{CV literature.}
Transfer learning is the most
adopted methodology across literature.
In particular, while early meta-learning literature proves
it is possible to learn even from single samples~\cite{snell2017neurips-prototnet}, 
more recent literature~\cite{chen2019icml-baseline,tian2020eccv-rethingkingfewshot}
suggests transfering from models created on large dataset (with many samples and many classes)
yields better performance, especially when combined with fine-tuning with episodic
training for the final target task, i.e., they mix
transfer learning with meta-learning.
\begin{table}[t]
\centering
\scriptsize
\caption{
Computer vision literature summary.
\label{tab:sota-cv}}
\begin{tabular}{
    @{}
    c
    r@{$\,\,$}
    l@{$\,\,$}
    c@{$\,\,$}
    c@{$\,\,$}
    c@{$\,\,$}
    @{}
}
& \multicolumn{2}{l}{\textbf{Approach}}  & \textbf{KnowDistil?} & \textbf{Classifier} & \textbf{Dist-based?}\\
\toprule
\multirow{6}{*}{\parbox[c]{3mm}{\bf \tiny Transf.\\Learn.}} &
\cite{chen2019icml-baseline} & $\mathtt{Baseline}$                              & \circleempty   & Linear  & \circleempty\\
& \cite{chen2019icml-baseline} & $\mathtt{Baseline}\text{++}$                   & \circleempty   & Cosine Sim. & \circlefull\\
& \cite{tian2020eccv-rethingkingfewshot} & $\mathtt{RFS\text{-}simple(LR)}$     & \circleempty   & Logistic Reg. & \circleempty\\
& \cite{tian2020eccv-rethingkingfewshot} & $\mathtt{RFS\text{-}distill(LR)}$    & \circlefull    & Logistic Reg. & \circleempty\\
& \cite{tian2020eccv-rethingkingfewshot} & $\mathtt{RFS\text{-}simple(NN)}$     & \circleempty   & Nearest Neighbor & \circlefull\\
& \cite{tian2020eccv-rethingkingfewshot} & $\mathtt{RFS\text{-}distill(NN)}$    & \circlefull    & Nearest Neighbor & \circlefull\\
\midrule
\multirow{3}{*}{\parbox[c]{3mm}{\bf \tiny Meta.\\Learn.}} &
\cite{snell2017neurips-prototnet} & \proto           & \circleempty    & Euclidean distance & \circlefull\\
& \cite{sung2018cvpr-relationnetwork} & \relation    & \circleempty    & MSE                & \circleempty\\
& \cite{finn2017icml-maml} & \maml                   & \circleempty    & Linear             & \circleempty\\
\midrule
\multirow{2}{*}{\parbox[c]{3mm}{\bf \tiny Contr.\\Learn.}} &
\cite{chen2020icml-simclr} & \simclr                   & \circleempty    & none    & \circlefull\\
& \cite{khosla2020neurips-supcon} & \supcon            & \circleempty    & none    & \circlefull\\
\bottomrule
\end{tabular}
\\
\raggedright
\end{table}

A prominent example is $\mathtt{Baseline}$~\cite{chen2019icml-baseline} 
which trains a \modelbase model using a large dataset (many classes and many samples)
and then fine-tune a target task (from a disjoint set of classes) via episodic training.
$\mathtt{Baseline\text{++}}$~\cite{chen2019icml-baseline} is a variant of the same approach where the parameters of the 
classifier are treated as \emph{class embedding}. More in details, both methods rely on a
fully connected layer $\textbf{W}$$\in$$\mathbb{R}^{d \times c}$ where $d$ and $c$ represent the dimension of the latent space vectors and  the number of target classes respectively.
Unlike $\mathtt{Baseline}$ that simply uses a standard cross-entropy classification loss,
$\mathtt{Baseline}\text{++}$ relies on a cosine distance-based loss between 
$\textbf{W}$=$[\textbf{w}_1 \cdots \textbf{w}_c]$ columns and 
latent space projections $z$ of input samples---each column $\textbf{w}_i$
``embeds'' a class.
Similar mechanisms power also $\mathtt{RFS\text{-}simple(LR)}$ and $\mathtt{RFS\text{-}simple(NN)}$~\cite{tian2020eccv-rethingkingfewshot} 
differing from the previous only by the classifier $f_\varphi(\cdot)$
being a logistic regression---a linear layer with sigmoid
activations rather than softmax---and a 1-nearest neighbour approach 
respectively---the model does not have a classifier layer but rather 
classifies based on proximity to pre-defined labeled projected samples.
$\mathtt{RFS}$ authors also introduced 
two other variants, namely \rfsdlr and \rfsdnn, based on 
\emph{self-distillation}, a special form of knowledge distillation~\cite{hinton2015distilling}.
In a nutshell, to reinforce models knowledge, the 
trained model is further fine-tuned by
replaying $\mathcal{D}_{train}$ and pairing
the classification loss with an additional term
comparing a smoother version (via temperature scaling)
of logits with respect to the fine-tuned logits---in
practice, the task becomes intentionally more complex
to further push the learning.
It follows that \rfsdlr and \rfsdnn use a 3-steps training: training the source model,
apply distillation and finally fine-tune the target tasks.

\begin{table}[t]
    \centering
    \scriptsize
    \caption{Traffic classification literature summary.}
    \label{tab:sota-tc}
    \begin{tabular}{
        @{}
        c@{$\,\,$}
        r@{$\,\,$}
        l@{$\,\,$}
        l@{$\,\,$}
        r@{$\,\,\,$}
        r@{$\,\,$}
        r@{$\,\,$}
        r@{$\,\,$}
        c@{$\,\,$}
        c@{}
    }
    & &  & & \bf Data & \multicolumn{2}{@{$\:$}c@{$\,\,$}}{\bf Classes} & &
    \\
     &
     \multicolumn{2}{l}{\bf Reference} &  
     \bf Approach &
     \bf Year &
     \emph{All} &
     \emph{Target} &
     \bf Shots &
     \bf Inp.Type &
     \bf Inters?
      \\
     \toprule
     \multirow{2}{*}{\parbox[c]{6mm}{\bf \tiny Transf.\\Learn.}} &
         \cite{rezaei2020icccn-multi-task-learning} & \emph{Razaei20}    & Multi-task learn.  & 18 & 5     & 5    & n.a.  & 3 PS  & \circlefull \\
         & \cite{sun2018-transfer-learning} & \emph{Sun18}                 & $\mathtt{TrAdaBoost}$         & 05 & 12    & 12   & n.a.  & -- FF & \circlefull \\
    \midrule
    \multirow{2}{*}{\parbox[t]{6mm}{\vspace{2.5em}\tiny \bf Meta\\Learn.}} &
         \cite{ouyang2021tc-fs-ids} & FS-IDS                             & \proto             & 14    & 8     & 3   & 1:10  & 26~FF & \circlesemi\\ 
         & \cite{rong2021ijcnn-umvd-fsl-unseen-malware} & UMVD-FSL         & \proto             & 17,20 & $\approx$76 & 5/20& 1/5   & 784~B & \circleempty \\ 
         & \cite{yu2020ieeeaccess} & \emph{Yu20}                           & \proto             & 09,15 & $\approx$8 & 2/5 & 50 & 44~FF & \circlefull \\ 
         & \cite{liang2022-toii-oics-vfsl} & OICS-VFSL                     & \proto             & 09,15 & $\approx$12 & 2/4/7 & 1:50 & ---FF & \circlefull \\
         & \cite{zheng2020www-rbrn} & RBRN                                 & \relation          & 12,16 & $\approx$15 & --- & --- & ---B & \circlefull 
         \\
         & \cite{xu2020tifs-fcnet} & FCNet                                 & \relation          & 12,17 & $\approx$10 & 2 & 5/10 & 200~B & \circlesemi 
         \\
         & \cite{zhao2022ieeecomlet-festic} & Festic                       & \relation          & 18,19 & $\approx$25 & 14 & 5:15 & 256~B & \circlefull 
         \\
         & \cite{feng2021-ifipnetworking-fcad} & FCAD                      & \maml              & 17    & 43 & 13 & 5:20 & 33FF+8PS & \circleempty 
         \\
    \midrule
    \multirow{1}{*}{\parbox[c]{6mm}{\bf \tiny Contr.\\Learn.}} &
         \cite{horiwicz2022imc-fsl-contrastive-learning} & \emph{Horiwicz22} & \simclr       & 18     & 5  & 5  & --- & FlowPic & \circlefull      
         \\
         \bottomrule
    \end{tabular}
    \\
\raggedright
\rule{0pt}{1em}
\hspace{-5pt}
\fontsize{6}{10}\selectfont{\textbf{InpType:} FF = flow features; B = payload bytes; PS = univariate packet time series; --- = unspecified}
\\
\fontsize{6}{10}\selectfont{\textbf{Inters?} target and base classes are \circleempty disjoint or either \circlefull completely or \circlesemi partially overlapped.}
         \\

\end{table}

\noindent \textbf{TC literature.}
Only \cite{rezaei2020icccn-multi-task-learning,sun2018-transfer-learning}
used transfer learning but none of the methods mentioned above.
Namely, \cite{rezaei2020icccn-multi-task-learning} pre-trained
an unsupervised multi-task model targeting flows duration and bandwidth
which was then transferred to a 5 classes task. Results
show that this transfer was under-performing with respect 
to training directly a single-task model. 
Instead, \cite{sun2018-transfer-learning} 
used TrAdaBoost~\cite{dai2007-tradaboost}, an ensemble ML method using reversed boosting,
considering a very old dataset.

\subsection{Meta-learning approaches}
\label{sec:meta_learning}

\noindent \textbf{CV literature.}
Computer vision literature is
ripe with meta-learning and FSL methods~\cite{wang2020survey-fsl}.
In this study we focus on a small selection of methods
based on their extreme popularity.

\vspace{2pt}
\noindent
\proto~\cite{snell2017neurips-prototnet} is the most well known metric-based meta-learning 
approach. \proto learns class prototype which
geometrically corresponds to the mean centroid of a class in the latent space.
Query samples are then classified based on their euclidean distance with respect to
class prototypes. The idea of class prototypes inspired different
meta-learning methods, including \basepp (i.e., class embedding 
are semantically equivalent to class prototypes).

\vspace{2pt}
\noindent
\relation~\cite{sung2018cvpr-relationnetwork} is another popular 
metric-based meta-learning method. While \proto uses a closed form
distance metric (i.e., euclidean distance), \relation introduces
the idea of ``meta-learning'' such distance. Specifically, the classifier $f_\varphi(\cdot)$ 
embodies a ``relation'' module trained to provide a similarity score
between support and query samples. Curiously, the classification loss
is based on Mean Squared Error (MSE) rather than softmax.

\vspace{2pt}
\noindent \maml~\cite{finn2017icml-maml} is the most popular optimization-based meta-learning approach.
Differently from \proto and \relation which optimize model parameters 
considering episodes in isolation,
\maml uses a two-nested loops process: the inner loop fine-tunes based on
each individual episode; the outer loop ``re-weights'' inner loop contributions
across episodes via a second order gradient of the classification loss.

\vspace{2pt}
\noindent \textbf{TC literature.}
As from Table~\ref{tab:sota-tc},
several studies successfully applied 
these methods on network traffic, mostly
targeting normal-vs-attack classification tasks 
in intrusion detection scenarios.
While we consider those classifiers as specific
forms of TC, we find \emph{most of these studies
violate the meta-learning principle of dis-joining train/val/test partitions}. 
Only \cite{rong2021ijcnn-umvd-fsl-unseen-malware,feng2021-ifipnetworking-fcad}
followed the expected protocol, while in all the other studies the 
partitions overlapped either perfectly (e.g.,
meta-train a binary classifier normal-vs-attack
and meta-test on the same classes) or partially
(e.g., normal and the same attack classes traffic belongs to both
meta-train and meta-test).

\subsection{Contrastive learning}

\noindent \textbf{CV literature.} Similarly to FSL,
also contrastive learning is a very active research area.
Across variants, \simclr~\cite{chen2020icml-simclr}
is the most popular method.
\supcon~\cite{khosla2020neurips-supcon} extends \simclr for a ``supervised'' setting by
simply considering as positive samples also all
other augmented version of samples belonging to 
the same class of the selected anchor---it moves from
self-similarity to class-similarity.

\vspace{2pt}
\noindent \textbf{TC literature.} We find only two previous
work using contrastive learning for TC.
In~\cite{horiwicz2022imc-fsl-contrastive-learning}, authors
applied contrastive learning
in a ``few-shot learning settings'' without episodic training.
More specifically, authors trained an unsupervised model
using \simclr and with a flowpic input representation---packet 
time-series are transformed
into images representing the evolution of traffic
over a time window; hence transformation is possible by
either manipulating the time series (e.g., time shift) or the related image (color jittering, occlusion, etc.)---and 
transferred it to the supervised classification task using 
just a few labeled samples. In~\cite{ziyi-networking22-cl-etc} instead, authors
apply BYOL~\cite{grill2020neurips-byol}---another
popular contrastive learning methods that does not 
rely on negative samples---to a similar problem setting 
as in \cite{horiwicz2022imc-fsl-contrastive-learning}.

Worth also of mention is \cite{rezaei2019ICDM-ucdavis} where authors
used augmentations similar to the one used in \cite{horiwicz2022imc-fsl-contrastive-learning},
yet no contrastive learning was applied.



\section{Research questions}
\label{sec:research-questions}

TC literature favors meta-learning 
with respect to transfer learning (Table~\ref{tab:sota-tc}).
Yet, computer vision literature suggests to pay
attention to the latter. 
Moreover, all previous TC literature on meta-learning
provides positive results, but
most of these studies violate
the meta-learning principle of dis-joining the classes in
$\mathcal{D}_{train}$, $\mathcal{D}_{val}$ and
$\mathcal{D}_{test}$.
Thus, we claim the need to re-assess these
methodologies under different settings.

\vspace{5pt}
\begin{quote}
\small
\it $\mathsf{Q1}$: 
Is transfer learning yielding better performance
compared to meta-learning?
Do the benefits, as observed in computer vision,
of transferring
from source models trained on many samples
and many classes apply to TC use-cases as well?
\end{quote}
\vspace{5pt}

FSL targets scenarios suffering from limited amount of
labeled samples. For instance, in TC this happens
when new applications are introduced. However,
differently from computer vision, we can also
rely on traditional ML models to address 
these small tasks. Conversely previous TC literature
focused mostly on DL methods adopting
image-like input representation, while we argue
that time series input is a more natural choice
for network monitoring.

\vspace{5pt}
\begin{quote}
\small
\it $\mathsf{Q2}$: How meta-learning performance compares
with common ML tree-based models when considering
as input time series of properties of the first $P$ packets of a flow? 
Do more shots improve performance?
\end{quote}
\vspace{5pt}

Next to transfer learning and meta-learning, also
contrastive learning is designed to facilitate better
representation learning. Since it has been shown 
to be successful for time series~\cite{yang2022-timeclr} outside
the TC domain, we ask

\vspace{5pt}
\begin{quote}
\small
\it $\mathsf{Q3}$: Does contrastive learning assist in creating
more general models also for TC? Is supervised contrastive learning
superior to its more traditional version?
\end{quote}
\vspace{5pt}




\section{Methodology}
\label{sec:methodology}



To address our research questions we relied on two publicly available
datasets and designed multiple modeling campaigns.

\subsection{Datasets}
\label{sec:datasets}

\noindent\textbf{\mirage}~\cite{mirage2019}
encompasses per-biflow traffic logs of $40$ Android apps 
collected at 
the ARCLAB laboratories of the University of Napoli Federico II.
It was collected by instrumenting $3$ Android devices 
used by $\approx$$300$ volunteers (students and researchers)
interacting with the selected apps for short sessions.
Each session resulted in a pcap file and an \texttt{strace}\footnote{https://man7.org/linux/man-pages/man1/strace.1.html} log 
mapping each socket to the corresponding Android package name.
Pcaps were then post-processed to obtain biflow logs by grouping all packets belonging to the same 5-tuple (srcIP, srcPort, dstIP, dstPort, L4proto) and extracting both aggregate metrics (e.g., total bytes, packets, etc.), per-packet time series (packet size, direction, TCP flags, etc.), raw packets payload bytes
(encoded as list of integer values) and mapping a ground-truth label by means of the \texttt{strace} logs.

\vspace{2pt}
\noindent\textbf{\appclassnet}~\cite{wang2022} 
is a commercial-grade dataset 
released by Huawei Technologies in 2022 and
gathered from real residential and enterprise networks.
The dataset encompasses the traffic of $500$ applications 
for a total of $10$M biflows each represented by a time series of packet-size and direction of the first $20$ packets and, most important, labeled by means of a commercial and proprietary DPI tool.
Prior to release, the dataset was anonymized to
remove \textit{privacy-sensitive} information 
(i.e., IP addresses, exact timing, protocol headers 
field values, and packet-payload) and \textit{business-sensitive} information
(i.e., applications labels are encoded as integer
values and raw time series values are anonymized).

\vspace{3pt}
\noindent \textbf{Dataset partitioning.}
Table~\ref{tab:datasets-summary} summarizes datasets properties. 
Defining the popularity of a class as its number of samples, 
the table reports $\rho$ measuring the datasets imbalance 
as the ratio between the most popular and least popular class.
As expected, the imbalance is severe; yet such condition
is rare in computer vision scenarios, especially in meta-learning settings---typical
FSL datasets like $\mathtt{Omniglot}$~\cite{lake2015human-omiglot}, 
$\mathtt{miniImageNet}$~\cite{vinyals2016matching-miniimagenet} and 
$\mathtt{CIFAR\text{-}FS}$~\cite{bertinetto2018metalearning-cifarfs}
have all $\rho$=$1$. Adhering to meta-learning protocols we 
partition the datasets by dis-joining train/val/test.
To do so, we use class popularity resulting in 
$\mathcal{D}_{train}$ containing the largest classes pool---imbalance here
still reflects network traffic scenarios and
the large availability of data allows to address ($\mathsf{Q1}$)---while $\mathcal{D}_{val}$
and $\mathcal{D}_{test}$ focus on unpopular classes---imbalance
here is reduced, better reflecting the typical meta-learning settings ($\mathsf{Q2}$).
We argue that such partitioning is preferable to both ($i$) artificially random under-sampling
to enforce a ``few-shot'' setting and ($ii$) randomly splitting classes
obtaining scenarios where a target class has many samples.
Conversely, we aim at target tasks with a naturally reduced number of samples.

\begin{table}[t]
\centering
\footnotesize
\caption{
Datasets summary.\label{tab:datasets-summary}}
\begin{tabular}{
    @{}
    r@{$\;\;$} 
    r@{$\;\;$}
    r@{$\;\;$}
    r@{$\;\;$}
    r@{$\;\;$}
    r@{$\;\;$} 
    r@{$\;\;$}
    r@{$\;\;$}
    r@{$\;\;$}
    r@{$\;\;$}
    r
    @{}
}
\toprule 
& \multicolumn{5}{c}{\mirage} & \multicolumn{5}{c}{\appclassnet}
\\
\cmidrule(r){2-6} \cmidrule{7-11}
\bf Data & 
\bf Num. &
\multicolumn{3}{c}{\bf Samples} &
\multirow{2}{*}{$\rho$} &
\bf Num. &
\multicolumn{3}{c}{\bf Samples} &
\multirow{2}{*}{$\rho$}
\\
\bf Partition & \bf Classes &\it All &\it Max & \it Min &
              & \bf Classes &\it All &\it Max & \it Min &
\\
\cmidrule(r){2-6} \cmidrule{7-11}
$\mathcal{D}_{train}$   & 24     & 82k   & 8.2k   & 1.3k  & 6.3  & 320  & 9.8M  & 1M  & 958  & 1,044 \\
$\mathcal{D}_{val}$     & 8      & 9.4k  & 1.3k   & 1.1k  & 1.2  & 80   & 60.5k & 956  & 579  & 1.6    \\
$\mathcal{D}_{test}$    & 8      & 5.1k  & 904    & 361   & 2.5  & 100  & 47.4k & 578  & 383  & 1.5    \\
\cmidrule(r){2-6} \cmidrule{7-11}
$\mathcal{D}_{all}$     & $40$   &  97k  & 8.2k  & 361  & 22.7    & 500    & 9.9M & $\approx$1M & $383$ & 2,611 \\
\bottomrule
\end{tabular}
\\
\raggedright
\rule{0pt}{1em}$\rho$ = ratio \emph{Max/Min} samples per class
\end{table}

\vspace{2pt}
\noindent \textbf{Input type.}
All models are created using packet time series as input ($\mathsf{Q2})$.
Specifically, for \mirage we consider 4 features (packet size, direction, inter-arrival time, and TCP window-size\footnote{For UDP traffic the time series is padded with 0.})
for the first 10 packets; for \appclassnet we consider 2 features (packet size and direction) for the first $20$ packets.\footnote{
The reason for the different time series length resides in properties of the datasets: \mirage 
contains many short leaved flows thus using 10 packets reduces padding.}

\subsection{DL methods}

\noindent \textbf{Approaches.}
We considered a total of 16 methods:
1 for monolithic training, 8 for transfer learning,
3 for meta-learning, and 4 for contrastive learning.
We use all methods reported in Table~\ref{tab:sota-cv}. 
However, for transfer learning, we modify original names
to better express their relationship and semantic: $\mathtt{Baseline\text{++}}$~$\rightarrow$~$\mathtt{Baseline(ClassEmb)}$,
$\mathtt{RFS\text{-}simple(LR)}$~$\rightarrow$~$\mathtt{Baseline(LR)}$ and $\mathtt{RFS\text{-}simple(NN)}$~$\rightarrow$ $\mathtt{Baseline(NN)}$.\footnote{We acknowledge that $\mathtt{Baseline}$ is 
a sub-optimal naming convention, but we kept it to
preserve the relationship with~\cite{chen2019icml-baseline}.}
Still on transfer learning, we added the two variants \basetl and \basepptl 
that fine-tune via monolithic training rather than episodic training.
For contrastive learning, we considered both unsupervised (\simclr) and
supervised (\supcon) variants ($\mathsf{Q3}$) considering models with and without $\mathtt{Baseline}$'s class
embedding (4 variants). We also randomly apply 4 transformations:
\emph{horizontal flip} reverses the order of packets (1st become last, etc.);
\emph{shuffle} randomly reorders packets;
\emph{tail-occlusion} masks the second half of an input time series with zeros;
\emph{gaussian noise} adds noise sampled from a normal distribution $\epsilon \sim \mathcal{N}(0,1)$ to each time series value.

\vspace{2pt}
\noindent \textbf{Architectures.}
We used CNN architectures based of
2d convolutional blocks (a convolution layer followed by a batch normalization layer and a ReLU activation):
$\mathtt{CNN\text{-}2}$ is inspired by \cite{lopez2017} and
has two CNN blocks with $32$ and $64$ filters respectively
followed by a fully connected layer of $200$ units (trunk with $529$k parameters);
$\mathtt{CNN\text{-}4}$ adds other two CNN blocks
and a fully connected layer of $500$ units (trunk $1.3$M parameters).

\vspace{2pt}
\noindent \textbf{Implementation.}
We complemented the Pytorch implementation provided by \cite{chen2019icml-baseline, tian2020eccv-rethingkingfewshot}
by adding monolithic training, traditional transfer learning, and contrastive learning methods.

\subsection{ML methods}
We compared all DL approaches
against a reference Random Forest (\RF) 
 provided by \texttt{sklearn}~\cite{sklearn} and fed with the concatenated packet time series
($\mathsf{Q2})$.

\subsection{Experimental scenarios}

We settled on scenarios intentionally 
designed to go beyond the traditional FSL settings where 
meta-training and meta-testing share the same $N$-ways, and 
number of shots is limited to $S$$\in$$\{1,5\}$.
Trained DL $\mathsf{M}_{source}$ models are used in
fixed representation (Sec.~\ref{sec:background-transfer-learning}).
We also experimented with 
different monolithic models to gather ``reference''
comparison performance.
Overall, the training campaigns run on multiple linux
servers equipped with multiple nVidia V100 GPUs.

\noindent \textbf{Monolithic models.}
When using monolithic training $\mathcal{D}_{val}$ is discarded\footnote{Alternatively we
could have merged $\mathcal{D}_{train}$ with $\mathcal{D}_{val}$, but transfer learning 
training would have enjoyed more classes than for meta-learning.}
and the actual training and validation set are obtained from 
a 9:1 partitioning of $\mathcal{D}_{train}$.
For \RF we used 100 estimators but varying \emph{max-dept} $\in$ $\{\text{unbounded}, 10, 30\}$.
For DL methods we used a batch size of 64 for \mirage
and 1,024 for \appclassnet with a static learning rate of 0.001.

\vspace{2pt}
\noindent \textbf{Episodic training.}
For episodic training, we varied the number of shots
$S$$\in$$\{5, 15, 50, 100, 200\}$
while keeping $Q$=$15$ query samples training
for $200$ epochs of $100$ episodes each. 
Following the reference 
implementation, the learning rate was (slightly) different between 
the three approaches:
\proto used a policy halving the learning rate every 10 epochs, while
\relation and \maml had a fixed learning rate of 0.001 and 0.0001 respectively.
We also experimented with varying both
training and testing ways (see Sec.~\ref{sec:heatmaps}).
The only exception is \maml which in the reference implementation
was not flexible enough to easily change models classifier
during meta-testing.\footnote{We found the same to be true
in other publicly available implementations of \maml; we believe
that this constraint can be alleviated but we leave it as future work.}

\vspace{2pt}
\noindent \textbf{Evaluation metric.}
We measured classification performance using \emph{balanced 
accuracy} defined as a weighted average of per-class accuracy
based on the inverse of the class popularity~\cite{balanced-accuracy}.


\begin{table}[t]
    \centering
    \scriptsize
\caption{Models reference performance.
}
\label{tab:models-reference}

\begin{tabular}{
    @{}
    r@{$\,$}
    c@{$\,\,$}
    c@{$\:\:$}
    c@{$\:\:$}
    c@{$\:\:$}
    r@{$\,$}
    l
    c@{$\:\:$}
    r@{$\,$}
    l
    @{}
    }
\toprule
& & & & \multicolumn{3}{c}{\mirage}
      & \multicolumn{3}{c}{\appclassnet}
\\
\cmidrule(r){2-4}
\cmidrule(r){5-7}
\cmidrule{8-10}
\bf Arch & & \bf Train & \bf Test & \bf Accuracy & \multicolumn{2}{c}{\bf Params.}
         & \bf Accuracy & \multicolumn{2}{c}{\bf Params.}\\
\cmidrule(r){2-4}
\cmidrule(r){5-7}
\cmidrule{8-10}
     \RF      &$(a)$          & \emph{all} & \emph{all} & $86.29$\tinytinyalternate{0.52} & $2.5$M    &/ $56$
                                                  & $66.73\pm0.12$      & $226$M  &/ $116$\\
     \RF      &$(b)$          & \emph{all} & 4 target   & $83.30$\tinytinyalternate{0.21} & $2.5$M    &/ $56$
                                                  &$55.95\pm0.63$      & $226$M  &/ $116$\\
     \RF      &$(c)$          & 4 target   & 4 target   & $96.84$\tinytinyalternate{0.09} & $26$k &/ $22$
                                                  & $95.88$\tinytinyalternate{0.15} & $30$k &/ $23$\\
     $\mathtt{RF\text{-}10}$ &$(a)$   & \emph{all} & \emph{all} & $57.64$\tinytinyalternate{0.70} & $94.1$k &/ $10$
                                                          & $8.66$\tinytinyalternate{0.08}      &$163$k      &/ $10$\\    
    $\mathtt{RF\text{-}30}$  &$(a)$   & \emph{all} & \emph{all}& $86.25$\tinytinyalternate{0.47} & $2.4$M &/ $30$
                                                         & $52.55$\tinytinyalternate{0.25}      &$90$M      &/ $30$ \\
\cmidrule(r){2-4}
\cmidrule(r){5-7}
\cmidrule{8-10}
     \CNNtwo       &$(a)$     & \emph{all} & \emph{all} & $79.69$\tinytinyalternate{0.06}  & \multirow{3}{*}{$529$k+} & $8$k
                                                  & $83.91$\tinytinyalternate{0.41}  & \multirow{3}{*}{$529$k+} & $101$k\\ 
     \CNNtwo       &$(b)$     & \emph{all} & 4 target   & $74.47$\tinytinyalternate{1.29}  & & $8$k
                                                  & $76.21$\tinytinyalternate{1.84}  & & $101$k\\ 
     \CNNtwo       &$(c)$     & 4 target   & 4 target   & $84.35$\tinytinyalternate{0.95}  & & $804$
                                                  & $91.26$\tinytinyalternate{0.93}  & & $804$\\
\cmidrule(r){2-4}
\cmidrule(r){5-7}
\cmidrule{8-10}
     \CNNfour      &$(a)$     & \emph{all} & \emph{all} & $79.83$\tinytinyalternate{0.10}  & \multirow{3}{*}{$1.3$M+} & $20$k
                                                  & $84.56$\tinytinyalternate{0.31}  & \multirow{3}{*}{$1.3$M+}  & $250$k\\ 
     \CNNfour      &$(b)$     & \emph{all} & 4 target   & $73.40$\tinytinyalternate{1.33}  & & $20$k
                                                  & $75.12$\tinytinyalternate{1.57}  & & $250$k\\
     \CNNfour      &$(c)$     & 4 target   & 4 target   & $84.87$\tinytinyalternate{0.71}  & & $2$k
                                                  & $91.93$\tinytinyalternate{2.14}  & & $2$k\\
\bottomrule
\end{tabular}
\\
For $\mathtt{RF\text{*}}$, \emph{\{total nodes\}} / \emph{\{avg depth\}}; For $\mathtt{CNN}\text{*}$, \emph{\{trunk\} + \{classifier\}} params.
\end{table}

\begin{table*}[t]
\centering
\scriptsize
\caption{
Comparing of transfer, meta- and conntrastive learning in $4$-way \tasktarget tasks. 
\label{tab:shots}
}
\begin{tabular}{
    @{}
    r@{$\;\;$}
    c@{$\;\;\;$}
    c@{$\;\;\;$}
    c@{$\;\;\;$}
    c@{$\;\;\;$}
    c@{$\;\;\;\;$}
    c@{$\;\;\;$}
    c@{$\;\;\;$}
    c@{$\;\;\;$}
    c@{$\;\;\;$}
    c
    @{}
}
\toprule
& \multicolumn{5}{c}{\mirage} 
& \multicolumn{5}{c}{\appclassnet}
\\
\cmidrule(r){2-6}
\cmidrule{7-11}
\bf Approach &   $5$-shots &   $15$-shots &   $50$-shots &   $100$-shots &   $200$-shots
             &   $5$-shots &   $15$-shots &   $50$-shots &   $100$-shots &   $200$-shots
\\
\cmidrule(r){2-6}
\cmidrule{7-11}
\base         &$60.24$\tinytiny{0.57}   &$72.62$\tinytiny{0.48}    &$82.26$\tinytiny{0.41}  &$86.09$\tinytiny{0.36}  &$88.72$\tinytiny{0.32}
              &$77.30$\tinytiny{0.65}   &$85.42$\tinytiny{0.50}    &$90.11$\tinytiny{0.38}  &$91.51$\tinytiny{0.38}  &$93.03$\tinytiny{0.32}
\\
\basepp       &$59.98$\tinytiny{0.54}   &$70.78$\tinytiny{0.49}    &$79.03$\tinytiny{0.42}  &$82.16$\tinytiny{0.41}  &\WORSTPERF$84.48$\tinytiny{0.38}
              &$76.54$\tinytiny{0.61}   &$84.35$\tinytiny{0.52}    &$89.16$\tinytiny{0.41}  &$91.39$\tinytiny{0.34}  &$92.30$\tinytiny{0.32}
\\
\rfsslr       &$65.50$\tinytiny{0.61}   &$75.28$\tinytiny{0.48}    &$82.40$\tinytiny{0.42}  &$85.87$\tinytiny{0.38}  &$87.99$\tinytiny{0.34}
              &$77.42$\tinytiny{0.65}   &$84.03$\tinytiny{0.53}    &$88.77$\tinytiny{0.41}  &$90.48$\tinytiny{0.39}  &\WORSTPERF$91.83$\tinytiny{0.34}
\\
\rfssnn       &$65.48$\tinytiny{0.56}   &$77.13$\tinytiny{0.48}    &$86.41$\tinytiny{0.36}  &$90.26$\tinytiny{0.30}  &\BESTPERF$92.84$\tinytiny{0.24}
              &$76.93$\tinytiny{0.61}   &$84.81$\tinytiny{0.51}    &$90.18$\tinytiny{0.39}  &$92.25$\tinytiny{0.33}  &$93.77$\tinytiny{0.29}
\\
\rfsdlr       &$64.62$\tinytiny{0.56}   &$75.14$\tinytiny{0.47}    &$82.91$\tinytiny{0.41}  &$86.10$\tinytiny{0.36}  &$88.46$\tinytiny{0.33}
              &$77.10$\tinytiny{0.65}   &$83.73$\tinytiny{0.52}    &$88.69$\tinytiny{0.42}  &$90.64$\tinytiny{0.38}  &$92.14$\tinytiny{0.34}
\\

\rfsdnn       &$64.85$\tinytiny{0.58}   &$77.37$\tinytiny{0.47}    &$86.78$\tinytiny{0.36}  &$90.10$\tinytiny{0.30}  &$92.51$\tinytiny{0.25}
              &$76.65$\tinytiny{0.66}   &$85.02$\tinytiny{0.50}    &$89.77$\tinytiny{0.40}  &$92.45$\tinytiny{0.34}  &\BESTPERF$93.83$\tinytiny{0.30}
\\

\cmidrule(r){2-6}
\cmidrule{7-11}

\maml         &$57.10$\tinytiny{0.58}   &$65.19$\tinytiny{0.48}    &$70.68$\tinytiny{0.44}  &$73.92$\tinytiny{0.44}  &\BESTPERF$73.97$\tinytiny{0.44}
              &$61.93$\tinytiny{0.71}   &$72.60$\tinytiny{0.66}    &$75.57$\tinytiny{0.60}  &$77.52$\tinytiny{0.57}  &$78.27$\tinytiny{0.55}
\\
\proto        &$62.62$\tinytiny{0.56}   &$67.07$\tinytiny{0.47}    &$69.93$\tinytiny{0.41}  &$70.72$\tinytiny{0.42}  &$72.09$\tinytiny{0.40}
              &$69.93$\tinytiny{0.74}   &$77.81$\tinytiny{0.63}    &$80.31$\tinytiny{0.51}  &$81.05$\tinytiny{0.52}  &\BESTPERF$81.94$\tinytiny{0.50}
\\
\relation     &$54.28$\tinytiny{0.58}   &$58.73$\tinytiny{0.50}    &$57.73$\tinytiny{0.51}  &$62.99$\tinytiny{0.45}  &\WORSTPERF$61.60$\tinytiny{0.47}
              &$68.65$\tinytiny{0.74}   &$72.22$\tinytiny{0.67}    &$77.24$\tinytiny{0.59}  &$78.54$\tinytiny{0.56}  &\WORSTPERF$75.09$\tinytiny{0.57}
\\
\cmidrule(r){2-6}
\cmidrule{7-11}
\basesimclr   &$63.97$\tinytiny{1.01} &$74.61$\tinytiny{0.81} &$79.26$\tinytiny{0.76} &$80.94$\tinytiny{0.58} &\WORSTPERF$81.52$\tinytiny{0.63} 
&$77.88$\tinytiny{2.05}   &$86.73$\tinytiny{1.43}    &$91.10$\tinytiny{1.16}  &$91.18$\tinytiny{1.15}  &\WORSTPERF$91.65$\tinytiny{1.11} 
\\
\basesupcon   &$64.72$\tinytiny{0.83}   &$77.62$\tinytiny{0.69}    &$86.55$\tinytiny{0.50}  &$90.07$\tinytiny{0.43}  &$91.00$\tinytiny{0.39} 
&$81.74$\tinytiny{1.07}   &$89.29$\tinytiny{0.82}    &$91.70$\tinytiny{0.64}  &$92.03$\tinytiny{0.60}  &$93.33$\tinytiny{0.51} 
\\
\baseppsimclr &$62.82$\tinytiny{0.98}   &$76.53$\tinytiny{0.83}    &$86.91$\tinytiny{0.57}  &$89.89$\tinytiny{0.46}  &$91.01$\tinytiny{0.43}
&$78.27$\tinytiny{2.17}   &$84.70$\tinytiny{1.84}    &$92.05$\tinytiny{1.08}  &$92.35$\tinytiny{0.98}  &$93.35$\tinytiny{0.92}
\\
\baseppsupcon &$66.42$\tinytiny{0.84}   &$77.69$\tinytiny{0.68}    &$87.01$\tinytiny{0.48}  &$90.32$\tinytiny{0.43}  &\BESTPERF$91.87$\tinytiny{0.37} 
&$81.05$\tinytiny{1.09}   &$89.63$\tinytiny{0.75}    &$93.75$\tinytiny{0.53}  &$95.18$\tinytiny{0.48}  &\BESTPERF$95.94$\tinytiny{0.44} 
\\
\bottomrule
\end{tabular}
\end{table*}


\section{Evaluation}
\label{sec:evaluation}

\subsection{Reference monolithic models}
We evaluated \RF and DL monolithic 
models in three scenarios: 
($a$) training and testing on all classes, ($b$) training on all classes
but testing a random set of 4 unpopular classes and ($c$) training and testing
on a random set of unpopular classes only.
We repeated the experiments 10 times, each with $5$-folds cross-validation
and $300$ random selection of 4 unpopular classes for ($b\text{-}c$).
Table~\ref{tab:models-reference} reports
the average balanced accuracy, $95th$ Confidence Intervals (CI)
and models size, i.e., number of nodes and depth
for {\RF}s and number of trunk and classifier parameters for CNNs.

\vspace{2pt}
\noindent \textbf{Random forests.}
Unbounded \RF models yield the best performance across all scenarios.
Yet, ($a\text{-}b$) models are ``unrealistic upper bound''
given their incredibly high complexity.
Specifically, pickle
serialization created files of $930$MB and $416$GB,
i.e., $422\times$ and $190$M$\times$ a \CNNtwo size for \mirage and \appclassnet
respectively.\footnote{Sizes do not account for gzip, zlib, etc., compression 
thus roughly reflect the memory required to load those models
for inference.} Reducing trees depth makes models leaner, but
still large in absolute terms---with \emph{max-depth}=$30$ we
obtain $880$MB and $21.1$GB for \mirage and \appclassnet
respectively---and aggressively reducing depth severely
affect performance.\footnote{We control trees size 
by means of max-depth for simplicity and we acknowledge that
also the estimators should be taken into account to 
make our assessment even more robust.}
Considering ($c$) instead, \RF models are a very lean option
as they yield the highest accuracy with small footprint ($\approx2$MB).

\vspace{2pt}
\noindent \textbf{CNNs.}
Doubling the depth of the CNN
provides $<$1\% performance improvement,
yet both \CNNtwo and \CNNfour have 
a small memory footprint ($2.2$MB and $5.4$MB respectively).

\vspace{2pt}
\noindent \textbf{Takeaways.} \emph{
On the one hand, while \RF reaps the performance by dynamically growing its
``architecture'', this can lead to large memory requirements.
Yet, for tasks with 4 classes with $\leq$$1,\!000$ samples \RF is still 
the best monolithic reference considering that DL models bounds their deployment to
($i$) GPU/TPU accelerators availability
and ($ii$) non trivial performance optimization.
These aspects have been overlooked by 
TC literature which prefers large architectures even
for relatively simple tasks---e.g., \cite{chen2017bigdata-seq2img} uses $7$M
parameters on a 5-classes task; \cite{wei2017icon-big-model}
uses a $3$M parameters on a 20-classes task.
This call for an in-depth investigation of DL architectures
taking into account a variety of datasets which is outside
the scope of this work. Conversely, we take the reverse
approach with respect to the literature and we rely on 
\CNNtwo for the explicit purpose
of investigating how much can be learned
with a relatively small architecture.
}

\subsection{Sensitivity to number of shots}
\label{sec: analysis_on_shots}

We continue the analysis considering 
transfer learning, meta-learning, and contrastive learning approaches
on one specific settings.
Recall that for transfer learning $\mathcal{T}_{source}$
contains all $320$ for \appclassnet (\emph{viz.} $24$ for \mirage)
(Sec.\ref{sec:background-transfer-learning} and Sec.\ref{sec:datasets}).
We train $\mathsf{M}_{source}$ models for 200 epochs.
For meta-learning, each epoch has 100 episodes
with $N$=$64$ ways for \appclassnet (\emph{viz.} $16$ for \mirage), 
with $Q$=$15$ queries and 
varying shots $S$$\in$$\{5, 15, 50, 100, 200\}$ for both datasets.
For all methods, $\mathcal{T}_{target}$ is based on episodes of
$N=4$ ways and $Q = 15$ queries with
the same configurations of $S$ shots as in training.
In each scenario we used a 5-fold cross validation for $\mathsf{M}_{source}$ models
fine-tuning $1,000$ $\mathsf{M}_{target}$ models, each
with 4 random classes from $\mathcal{D}_{test}$. Table~\ref{tab:shots}
collects the average balanced accuracy and
related $95th$ CI; we further visually pinpoints the best (green) and worst (red)
method for each category when using 200 shots.

\vspace{2pt}
\noindent \textbf{Meta-learning.}
First of all, 
we observe a sharp benefit when increasing the number of shots ($\mathsf{Q2}$).
Recall that the typical setup in computer vision literature
limits $S \in \{1, 5\}$ shots, and also TC literature related 
to FSL adopts constrained settings (see Table~\ref{tab:sota-tc}).\footnote{
Except $\mathtt{ISCX\text{-}VPN\text{-}nonVPN}$ and $\mathtt{USTC\text{-}TFC}$,
popular datasets in TC (Table~\ref{tab:sota-tc}) enjoy $>$$1,000$ avg. samples/class 
(even when filtering flows with $>$$10$ packets), but lack classes variety, thus more shots are viable.}
Those regimes are the worst performing in our analysis
with models trained on \mirage suffering a $-7\%$ performance gap
compared to \appclassnet models.
In other words, 
a larger variety of classes/samples is beneficial, yet
meta-training seems affected by 
(not obvious) ``bottlenecks''.

When increasing the shots, beside the larger labeling effort,
two more subtle effects start to appear.
First, an increased computational resource cost---to go beyond 100 shots we had to
split training across multiple GPUs, not for the models themselves,
but due to the resource required to track gradients given the increased episode size.
Second, the higher the number of shots, the more
an episode resembles a ``bloated'' micro-batch of traditional
monolithic training---an episode is deprived of its ``synthetic random task'' nature (see Sec.\ref{sec:background-meta-learning}).
Overall, and differently from previous literature (see Table~\ref{tab:sota-tc}),
meta-learning methods fail to obtain solid generalization ($\mathsf{Q2}$).

\vspace{2pt}
\noindent \textbf{Episodic transfer learning.}
Conversely, transfer learning methods are the best performing ($\mathsf{Q1}$).
In particular, on \appclassnet all methods are within a $\pm1\%$ gap, 
while for \mirage we observe $\pm8.36\%$ between the best and worst performing methods with 200 shots. 
Recall that most of these methods differ mostly for the classifier $f_\varphi(\cdot)$.
In particular, on \appclassnet we can rank their performance
as
\emph{logistic regression $<$ 
class embedding $<$ linear layer $<$ nearest neighbor}.
However, on \mirage class embedding is the worst 
while nearest neighbor has $+4\%$ gap compared to
the 2nd performing classifier (logistic regression).
This hints that the latent space representation learned on \mirage 
is worse than the one learned on \appclassnet despite
the very different task complexity (32-vs-400 classes)---a 
nearest neighbor classifier is more flexible than
a linear one, which possibly justifies the better performance
on \mirage if classes are not well separated; 
yet, a nearest neighbour classifier 
needs to carry training data with the model
in order to have labeled ``anchors'' to use for the classification.
Unfortunately, investigating the latent space by means
of silhouette score and similar clustering metrics
did not provide conclusive answers about the discrepancies
between the datasets.
Lastly, while distillation can provide benefits on average,
the difference with respect to other methods falls within
the CI ranges, but we cannot completely discard those methods 
which were effective in computer vision literature~\cite{tian2020eccv-rethingkingfewshot}.

\vspace{2pt}
\noindent \textbf{Contrastive learning.}
As suspected, by leveraging label information, supervised contrastive learning outperforms
the traditional version---e.g., up to $+10.35\%$
and $3.98\%$ on \mirage and \appclassnet respectively ($\mathsf{Q3}$).
Moreover, while class embedding where ineffective
for transfer learning, they are beneficial for 
contrastive learning. However, the negative side
of those techniques is their computational cost:
each $\mathcal{T}_{source}$ model took $\approx$1 day of training!
By investigating the code, we deemed those costs to
the randomized data augmentation.
Thus, we believe that proper code refactoring can
further reap contrastive learning benefits.

\begin{table}[t]
    \centering
    \footnotesize
    \caption{Transfer learning without episodic training.}
    \label{tab:vanilla-transfer}
    \begin{tabular}{
        @{}
        r@{$\:\,$}
        c@{$\:\,$}
        c@{$\:\,$}
        c@{}
    }
    \toprule
    \bf Approach 
    &\fontsize{7.5}{10}\selectfont{\mirage}
    &\fontsize{7.5}{10}\selectfont{\appclassnet}
    &\fontsize{7.5}{10}\selectfont{\appclassnet$\rightarrow$\mirage}\\
    \cmidrule(r){2-3}
    \cmidrule{4-4}
    \basetl       & $86.62$\tinytiny{1.15}    & $95.48$\tinytiny{0.78}    & $78.69$\tinytiny{0.41}\\
    \basepptl     & $82.85$\tinytiny{1.54}    & $94.72$\tinytiny{0.85}    & $71.13$\tinytiny{0.45}\\
    \bottomrule
    \end{tabular}
\end{table}

\vspace{2pt}
\noindent \textbf{Transfer learning without episodes.}
To complete the benchmark, Table~\ref{tab:vanilla-transfer}
reports the performance when using transfer learning without
episodic training. We observe
that for \mirage episodic training is more beneficial than
for \appclassnet, even when compared against supervised
contrastive learning. 
As before, we suspect these discrepancies relates 
to differences in the latent space geometry 
which are not so obvious to extrapolate.

Table~\ref{tab:vanilla-transfer} also shows that
using \appclassnet as source model 
to fine-tune task related to \mirage 
yielded poor performance.
Recall that \appclassnet underwent an
anonymization process modifying raw features values.
Moreover, \mirage relates to mobile apps traffic while
the (private) dataset from which
\appclassnet was created relates to wired networks.
Based on these observations, we find quite remarkable
that this transfer still outperforms 
meta-learning methods for \mirage.

\vspace{2pt}
\noindent \textbf{Takeaways.}
\emph{
Differently from previous literature, the analysis
show poor performance for meta-learning, and suggest
to focus on transfer learning and contrastive learning ($\mathsf{Q1\text{-}2}$).
The latter of the two poses interesting engineering
challenges and the positive results we gathered
seem corroborating the idea of relying on self-supervision
to further boost representation learning~\cite{chen2020icml-simclr}.
Notice also how the best DL method is on-par (or close to)
the reference \RF on 4 classes tasks---this is a positive
signal of more general representations ($\mathsf{Q1}$).
}

\subsection{Sensitivity to the number of ways}
\label{sec:heatmaps}

\begin{figure}[t]
\centering

\includegraphics[
trim=10 800 10 810, clip,
width=\columnwidth
]{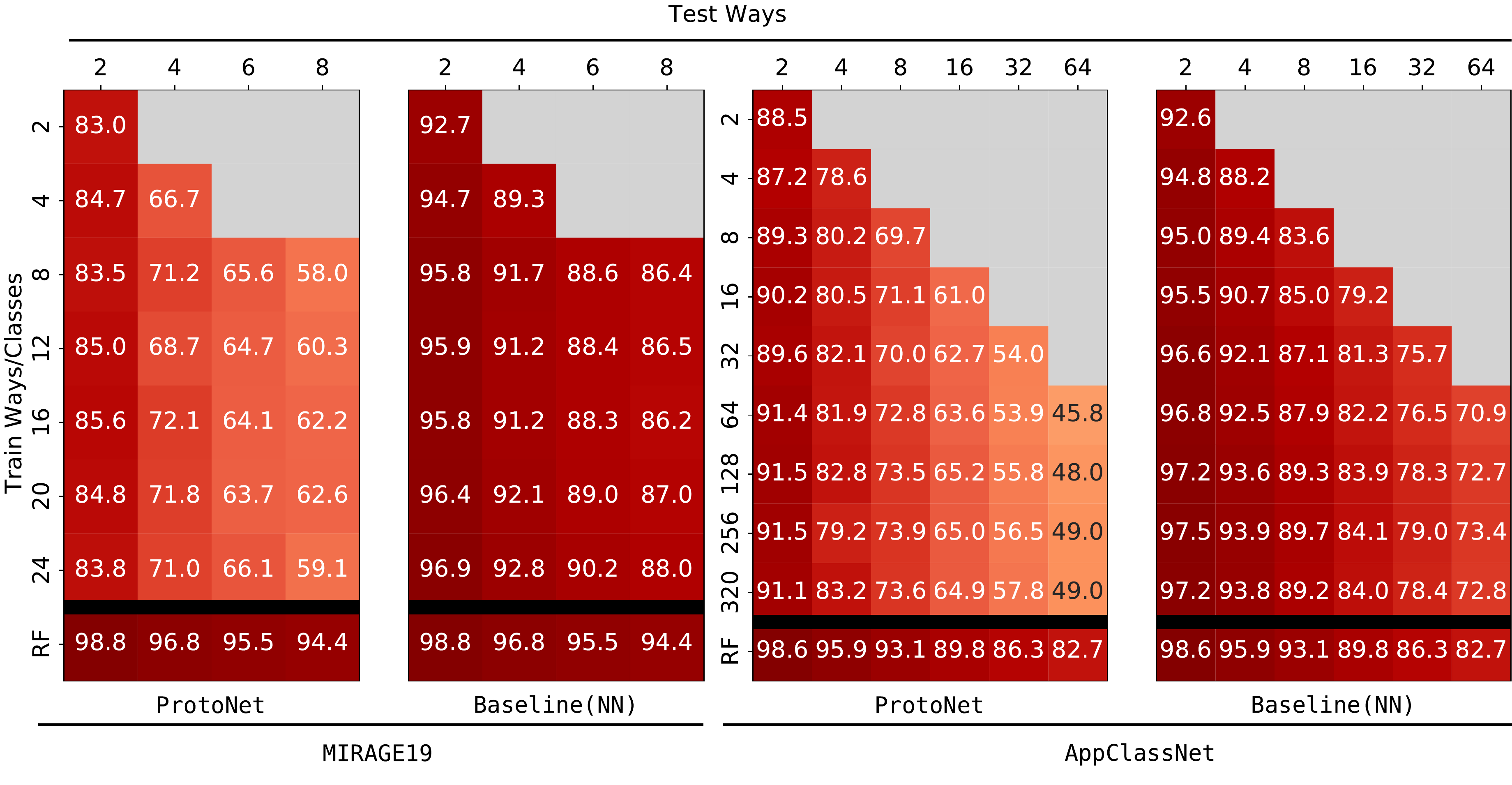}

\caption{
Sensitivity to train and test ways ($S=200$, $Q=15$).
}
\label{fig:heatmaps}
\end{figure}

Meta-learning methods evaluated in Table~\ref{tab:shots}
were trained with a smaller $N$-way that for transfer learning.
If on the one hand, each episode is less complex than
for transfer learning, it is possible that 
transfer learning methods enjoy a more ``regularized'' latent space 
due to the higher number of classes pushing for more 
knowledge extraction.
Thus, it is natural to wonder if the performance gap
is due to the experimental settings.
To investigate this, we run a second campaign
varying both number of training and testing ways.
For this analysis, we focused only on \proto and \rfssnn
as representative methods.
We fixed $S=200$ shots and $Q=15$ queries while varying the number of train ways $\{2,4,8,12,16,20,24\}$
and test ways $\{2,4,8\}$ for \mirage ($\{2,4,8,16,32,64,128,256,320\}$ and $\{2,4,8,16,32,64\}$ for \appclassnet).
As before, we trained the models using $200$ epochs of $100$ episodes each.
Fig.~\ref{fig:heatmaps} heatmaps report the average balanced accuracy across $1,\!000$ test episodes
(CI are in line with previous experiments, hence not reported for brevity) 
as well as the reference \RF performance. 

Starting from columns values, performance improve
when increasing the train ways---e.g., $+4.6\%$ for \rfssnn on \appclassnet 
when moving from 2 to 320 classes (160$\times$ more classes), 
yet only $+0.4\%$ when moving from 64 to 320 (5$\times$ more classes).
The same is true for \mirage too but with milder benefits.
The trend is reversed when considering rows values---e.g.,
on \mirage, $-8/9\%$ performance drop for \proto ($-2/3\%$ for \rfssnn)
every time we double $\mathcal{T}_{target}$ 
number of classes.
Overall, all models are under-performing compared to \RF models with unbounded depth.
Yet, also \RF performance decreases with similar
gaps to DL when doubling the target task size.

\vspace{2pt}
\noindent\textbf{Takeaways.} \emph{
The analysis suggests that better generalization can 
indeed be achieved when using a larger pool of classes ($\mathsf{Q1}$).
However, this is just a first step in the right direction.
Recall indeed that
we used fixed representation according to traditional
evaluation protocol for representation learning, and
we intentionally fixed models architecture too---our results are likely lower bounds.
}

\section{Discussion and Conclusion}
\label{sec:conclusion}

In this work, we compared transfer, meta-
and contrastive learning against traditional
ML tree-based models and DL monolithic training
using two publicly available datasets with
larger classes variety than in previous TC literature.

On the one hand, tree-based models achieve
the highest performance. Although
their size explodes for complex tasks,
when dealing with $\leq\!\!\!10$ classes  
(and an average of $1,\!000$ samples/class), they 
are still the most practical solution ($\mathsf{Q}2$).
On the other hand, compared with tree-based models, 
DL models handle more complex tasks, and
their performance is almost on-par with tree-base models also for small tasks
when using transfer learning methods---to the best
of our knowledge, this has been overlooked in
previous TC literature.

More important, results show that a large classes variety fosters
reuse of DL models ($\mathsf{Q1}$). In particular,
and differently from previous TC literature,
our results show that meta-learning methods
are the worst performing methods ($\mathsf{Q2})$, while 
transfer learning based on episodic fine-tuning
is a better option---training with many samples (and many classes)
is the best option to create $\mathsf{M}_{source}$ models;
these can then be used to fine-tune $\mathsf{M}_{target}$
models using $100$s samples. Moreover, data augmentations 
and supervised contrastive learning 
yield the best DL models overall ($\mathsf{Q3}$).

We acknowledge also some limitations. For instance,
we follow the traditional protocol of testing
representations via fixed representation ($\mathsf{M_{source}}$ is 
frozen when transferred), 
and we also adopt on relatively small architecture compared to TC literature,
i.e., our results are likely lower bounds. But, even in this
constrained setting, supervised contrastive learning offers
performance almost on-par with RandomForest
(when training tasks with 4 classes) ($\mathsf{Q3}$). We firmly believe that
performance can improve via further optimizations (e.g., different architectures, 
input type, training methodology, and data augmentation policies). 
However, with this work, we aim to spark
a ``mind shift'' in the TC research community to depart from simple
tasks ($\approx$$10$ classes), and refocus on
more challenging settings to push the learning. 
In particular, next to \mirage and \appclassnet, new large datasets
have been recently released~\cite{luxemburk2023-cesnet-tls,LUXEMBURK2023-cesnet-quic}
and more are likely to come: now we call to the research community to 
take them into action.


\footnotesize
\bibliographystyle{IEEEtranN}
\bibliography{bibliography.bib}

\end{document}